# Generation and Detection of Sign Language Deepfakes - A Linguistic and Visual Analysis

Shahzeb Naeem, *Member, IEEE,*
Muhammad Riyyan Khan, *Member, IEEE,* Usman Tariq, *Member, IEEE,*
Abhinav Dhall, *Member, IEEE,* Carlos Ivan Colon, and Hasan Al Nashash, *Senior Member, IEEE*

*Abstract*—**This research explores the positive application of deepfake technology for upper body generation, specifically sign language for the Deaf and Hard of Hearing (DHoH) community. Given the complexity of sign language and the scarcity of experts, the generated videos are vetted by a sign language expert for accuracy. We construct a reliable deepfake dataset, evaluating its technical and visual credibility using computer vision and natural language processing models. The dataset, consisting of over 1200 videos featuring both seen and unseen individuals, is also used to detect deepfake videos targeting vulnerable individuals. Expert annotations confirm that the generated videos are comparable to real sign language content. Linguistic analysis, using textual similarity scores and interpreter evaluations, shows that the interpretation of generated videos is at least 90% similar to authentic sign language. Visual analysis demonstrates that convincingly realistic deepfakes can be produced, even for new subjects. Using a pose/style transfer model, we pay close attention to detail, ensuring hand movements are accurate and align with the driving video. We also apply machine learning algorithms to establish a baseline for deepfake detection on this dataset, contributing to the detection of fraudulent sign language videos.**

*Index Terms*—**Human Analysis, Human Perception, Human Generation, Computer Vision, Human Data**

## I. Introduction

In recent times, deepfakes have garnered significant attention and witnessed remarkable progress, particularly in the context of facial manipulation. This has resulted in the focus being predominantly on facial aspects, leaving the exploration of AI (Artificial Intelligence) based human generation for the entire body aside. Techniques such as 3D facial modeling [1], computer vision based graphics rendering [2], and GAN-based face image synthesis [3] have been commonly employed in this regard. Notable approaches include facial expression transfer (facial reenactment), exemplified by Face2Face [4], and face swapping as in SimSwap [5].

Therefore, the question arises as to whether we can go beyond the facial deepfakes to a credible level. In addition, we ask if there is a benefit in doing so. Deepfakes have not proven to have the best of societal impacts as of yet either. However, our research aims to extend the deepfakes beyond just the facial elements, aiming to create fake videos that encompass the upper body as well, while addressing elements such as the hands and fingers for sign language production. This presents significant challenges, especially considering the complexities involved in generating credible videos with swift hand and body movements. These movements are a dominant part of sign language communication.

While existing work on half or full-body deepfakes has predominantly focused on images, our work bridges the gap by exploring the progression from facial to full-body deepfakes in sign language videos. An early method, "Pose Guided Image Generation" [6], emphasized the pose of a person and entire body features rather than just facial elements, marking a pivotal shift in focus. Subsequent work has witnessed substantial improvements and a shift in human perception for such works as models have become more sophisticated and efficient [7].

The transition from image-based to video-based pose transfer poses new challenges. "Liquid Warping GAN" [8] in 2017 offered a glimpse of deepfake videos encompassing the entire body, but the results were not satisfactory. Over the years, video-to-video and image-to-video generation faced challenges due to temporal context, blurry frames, and the handling of temporal aspects without compromising visual features. Computational costs for training or running such computer vision models, coupled with the difficulty in obtaining labeled human data for video generation, further complicate the scenario.

Despite advancements in diffusion models [9], recent works often focus on uncontrolled image and video human generation, frequently using text to control the output [10]. Many existing models use the supervised learning approach. Therefore, they are trained on heavily labeled datasets tailored for specific applications, making their adaptation to new datasets challenging. This entails extensive labeling, 2D and 3D pose extractions, and potentially incorporating text. Our research distinctively utilizes a source image using a driving video – allowing any person to present sign language in a video that is both visually credible and linguistically sound.

We propose an unsupervised model inspired by "Motion Representations for Articulated Animation" [11]. This model dynamically extracts key points during training or testing, eliminating the need for pre-labeled pose data while addressing the limitations of supervised approaches. Despite extensive research on sign language recognition and translation, there has been a noticeable gap in sign language production. Existing

Shahzeb Naeem, Muhammad Riyyan Khan, Usman Tariq and Hasan Al Nashash are with the Department of Electrical Engineering, College of Engineering, American University of Sharjah, Sharjah, 26666 UAE e-mails: (b00080174@aus.edu, b00096999@aus.edu, utariq@aus.edu, and hnashash@aus.edu).

Abhinav Dhall is with Data Science and AI Department, Monash University, Victoria, Australia e-mail: (abhinav.dhall@monash.edu).

Carlos Ivan Colon is with College of Applied Health Sciences, University of Illinois Urbana - Champaign, 61801, USA e-mail: (colon17@illinois.edu).

works often also rely on pose stick Figures diverging from our approach. Most of the work in this domain has been led by Ben Saunders [12], [13].

In a nutshell, our goal here is to make a pioneering contribution to accelerate work in the sign language production domain and create videos that are visually believable and technically & linguistically credible to the human perception. Henceforth, we do test the authenticity of our videos from both human analysis along with a visual and linguistic perspective through machine learning and a sign language expert. To the authors' best knowledge, this is one of the first works in this regard.

There are a plethora of spoken languages and over 300 sign languages [14]. It is also estimated that 1 in every 10 people will have some sort of hearing loss by 2050 [15] which would comprise a significant portion of the human population. Deepfakes pose new challenges such as fake news, rumour propagation, and several ethical concerns. Due to these reasons there have been several algorithms that target deepfake detection [16], [17], [18]. However, none of the aforementioned works focused on sign-language deepfake detection. This is also a novel contribution of our work. We create a dataset using our approach and present the pioneering work to detect deepfake sign-language videos. In addition, the presented work can be used in several other applications. The applications include, but are not limited to, content creation to deliver messages to the DHoH community with a celebrity who does not know sign language, choosing a sign language interpreter of choice, creating sign language videos with a more relatable person depending upon the geographical location, and so on.

The rest of the work is organized as follows. We first look at the types of style/pose transfer, the existing sign language datasets and the challenges related to them. After that, we delve into the pre-processing of data, the methodology behind the model being used to create the deepfakes and the post-processing of the output videos. Then we will explore the techniques related to analyzing the said sign language deepfakes and, then finally, we will present the results of those analyses.

## II. Types of Pose/Style Transfer

As generating sign language deepfakes involves a style/pose transfer problem on human data [19], [20], our emphasis lies in the dynamics of the various moving parts within one video frame compared to another. Both frames, constituting the source image and the driving video frame, can fall into three main categories with respect to one another. These frames may differ in various aspects, including clothing, accessories, size, and overall appearance of the person.

Let's explore the three types of pose and/or style transfer in our work:

1) The same person in the source image and driving video with different appearances
2) A different person in the source image and driving video with similar appearances
3) A different person in the source image and driving video with different appearances

Accommodating all three types becomes increasingly challenging as we move down the list. Consequently, we aim to create a deepfake dataset incorporating elements from all these three types. Our aim is to preserve or generate hands and facial expressions, ensuring that each sign language deepfake video is not only believable visually, but also interpretable in real life.

## III. Sign Language Dataset and Related Challenges

Sign language datasets have been collected since the advent of AI. The BSL (British Sign Language) Corpus [21] and SIGNUM [22] are continuous datasets (consisting of videos with multiple sign language expressions). Other datasets also exist but are non-continuous (consist of videos with isolated signs). The latter does not offer much benefit, as they lack more temporal information. At the same time, continuous datasets usually have the disadvantage of not being large enough to move towards sign language production successfully. Note that, these have been utilized for sign language recognition and translation with varying degrees of success. However, things started changing with the WLASL (World-Level American Sign Language) dataset [23], [24], which was one of the first attempts for compiling a comprehensive sign language dataset. Recently, large scale multimodal dataset for continuous American Sign Language, "How2Sign", was introduced [25]. Our work is based on this dataset.

### A. Challenges when choosing Sign Language Datasets

Many challenges arise when one takes a deeper look into the current datasets, especially relevant ones, when producing sign language videos. These include:

1) Lack of size (few videos)
2) Too much variability (background, environment, subject positioning, posture, or angle)
3) Poor lighting (dim or very lit surroundings)
4) Poor quality (not high definition)

In essence, when talking about training any AI model, things need to be streamlined, continuous, and without discrepancies. The lack of a sheer number of videos, labelled data, and consistency is a challenge that limits us severely. The better the dataset, the better the performance of a model. The poor visual aspects of several datasets mean that facial expressions, hands, and specifically fingers take a great hit in clarity, while generation.

### B. Our Chosen Sign Language Dataset

In terms of sheer size, one dataset stands above the rest. The "How2Sign" dataset [25], [26], although not consisting of many subjects (6 in total), has a substantial number of videos and covers a plethora of sign language expressions. Additionally, the videos are in high definition and maintain great consistency in terms of background and the frontal facing view of the subjects from a visual, human perception. The dataset has around 30,000 videos, making it very suitable for training a deepfake generation model, but the lack of enough subjects does create limitations. Henceforth, our approach also aims to address this deficiency in the created dataset for deepfake videos.

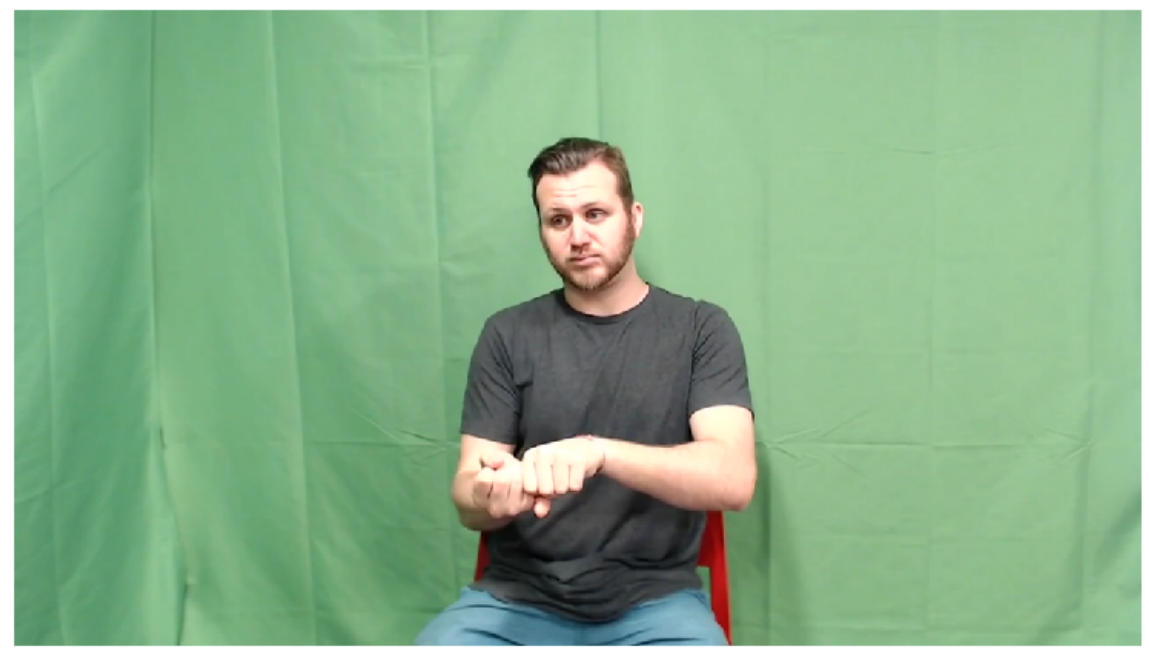

Fig. 1. A video frame of a subject before dynamic cropping

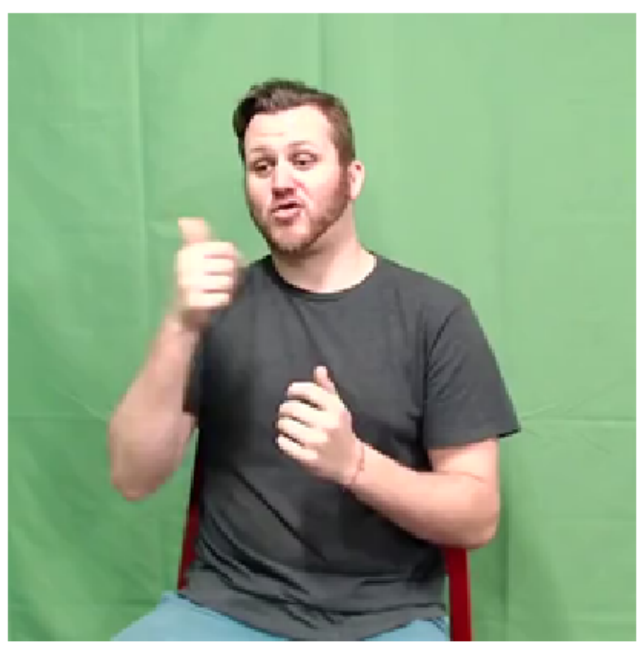

Fig. 2. A video frame of the same subject after dynamic cropping

## IV. DATA PRE-PROCESSING

Ensuring the prominence of key body parts, such as fingers, hands, and faces, is crucial for effective model training, especially when dealing with the relatively small and non-significant appearance of these elements within the frame of our chosen dataset. There is also a lack of consistency in that not all subjects are sitting in the same place with respect to the entire frame. Moreover, some are shorter in height, and some are thinner than others.

To bring the required uniformity to the videos we introduced dynamic cropping. It leveraged RetinaFace [27], a face detector, to facilitate the extraction of facial positions in each frame. It involved the incorporation of a variable scale factor, adapting to the facial size of each subject. Minor adjustments were made for individual subjects, tailoring the cropped window to ensure uniformity. The resulting cropped videos were standardized to a square format and downsized to 384x384 resolution. A comparison of before and after can be seen in Figures 1 and 2.

A total of approximately 500 videos were cropped for model training for deepfake generation, with an additional 100 videos reserved for evaluating the model. This balanced approach was appropriate considering the available resources and model complexity.

## V. METHODOLOGY

Our approach leverages a modified version of the First Order Motion Model (FOMM) [11], [28] for image animation. The enhancements we introduced facilitate precise hand accuracy and intricate detailing, surpassing expectations in applications of this nature in human body generation. Image animation typically involves making an entire image appear to move cohesively, rather than focusing on smaller isolated parts such as the fingers. In our application, fingers are of utmost importance.

Our model is fundamentally divided into two segments: Motion Estimation and Image Generation. The motion estimation phase involves key point extraction without the need for pre-existing key points before training, making it an unsupervised method.

Firstly, we extract coarse or sparse motions (flow vectors for features like edges/corners) from the source and driving frames. We employ an encoder-decoder key point predictor network (an autoencoder model with a modified U-Net [29] architecture). The output of this autoencoder is used to calculate the affine transformations based on a common reference frame. This provides us with the motion between all parts of the source and driving frames. Essentially, we end up having K heat maps in total that depict the regions of motion between the predicted moving parts/key points of the source and driving frames. The number of regions is fed to the model before training. In theory, the greater the number of regions, the more the intricate details (fingers for example) of a frame will be considered as separate moving parts. This would result in better motion prediction. Once the heat maps have been produced based on the affine transformations, soft argmax is used to estimate the translation component of the affine transformations, while PCA (Principal Component Analysis) based on SVD (Singular Value Decomposition) is employed for the computation of other transformations.

Secondly, dense motions (optical flow vectors for each pixel) between the source frame and driving frame are computed based on the previously obtained coarse motions. A pixel-wise flow predictor model, another modified U-Net [29] autoencoder selects the necessary coarse motions. Like before, K heat maps are generated, one per region of motion and another extra one for the background. Soft argmax is applied yet again in a pixel-wise fashion, and the optical flow between frames is computed by multiplying coarse heatmaps with dense heatmaps and adding their result to the background dense heatmap.

Thirdly, a confidence map is computed using the same autoencoder network used for the pixel-wise flow predictor, essential for handling missing parts in the source image (image inpainting). With both confidence map and optical flow field for each image (as seen in Figure 3) having been computed, the source image is passed through an encoder, then warped using the previously computed optical flow map, and finally multiplied by the previously computed confidence map. Finally, this image is passed through a decoder which reconstructs it into the driving frame as desired. The encoder-decoder structure of this part follows the Johnson Architecture, with deformable skip connections similar to the Monkey-Net [30] architecture.

The encoder-decoder and pixel-wise flow generator models are distinct modified versions of the U-Net architecture, while image generation follows a structure similar to the Johnson [31] architecture.

To address issues like lack of clarity in hands and occasional wrong predictions of motion, we increased the number of regions or the key points to be looked for by the model to 50. This adjustment allows the model to track more moving parts, emphasizing fingers and facial features, resulting in improved predictions for hands and facial expressions. As shown in Figure 4 on the TED dataset [32], our model produces clearer and more accurately translated hands in the output frame compared to the stock model as in Figure 3.

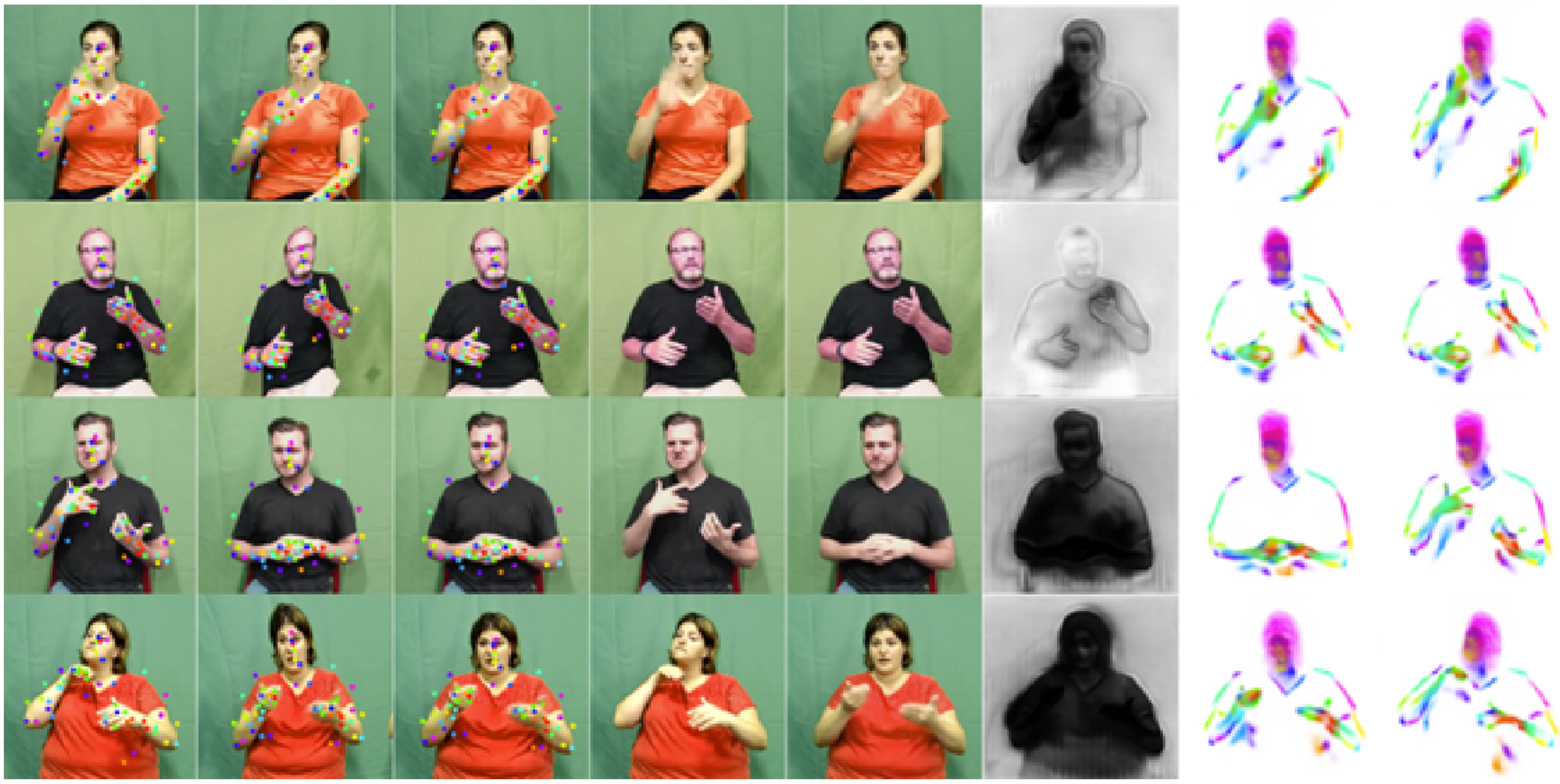

Fig. 3. An example of the generated confidence maps, the optical flow maps, and key-points.

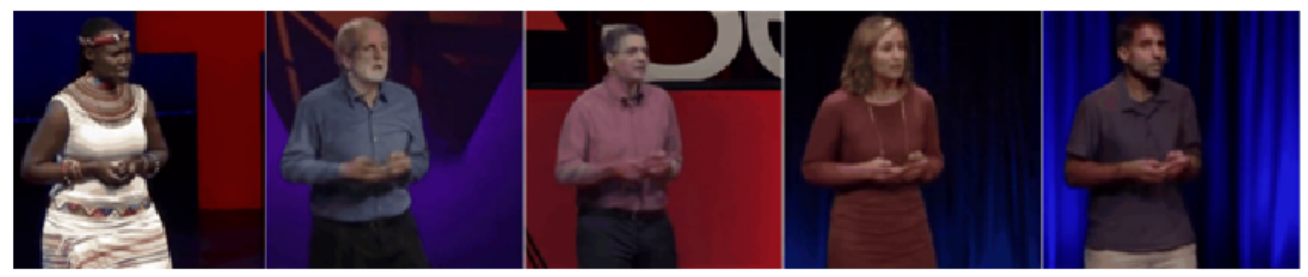

Fig. 4. A frame from the driving video is shown as the first image on the left. The others are outputs of the stock model on the TED dataset. One can note the hands are not well generated from the stock model, warranting further changes to the model.

Additionally, we aimed for enhanced clarity and improved learning by altering the VGG perceptual loss function [31]. We replaced the previous L1 loss [33] with the Charbonnier loss [34], while model training.

The rationale behind this change is that the L1 loss [33], measuring the absolute error between two values (pixels in our case), tends to sharpen the entire image, avoiding the squaring effect seen in L2 loss [33]. The L2 loss [35], a mean squared error loss, tends to make the image blurrier and is more affected by larger anomalies or outliers. Hence, to strike a balance between the two, we adopted the Charbonnier loss [34], which maintains image clarity while accounting for anomalies or outliers to a degree determined by us through control parameters. This novel addition, helped with generating better videos.

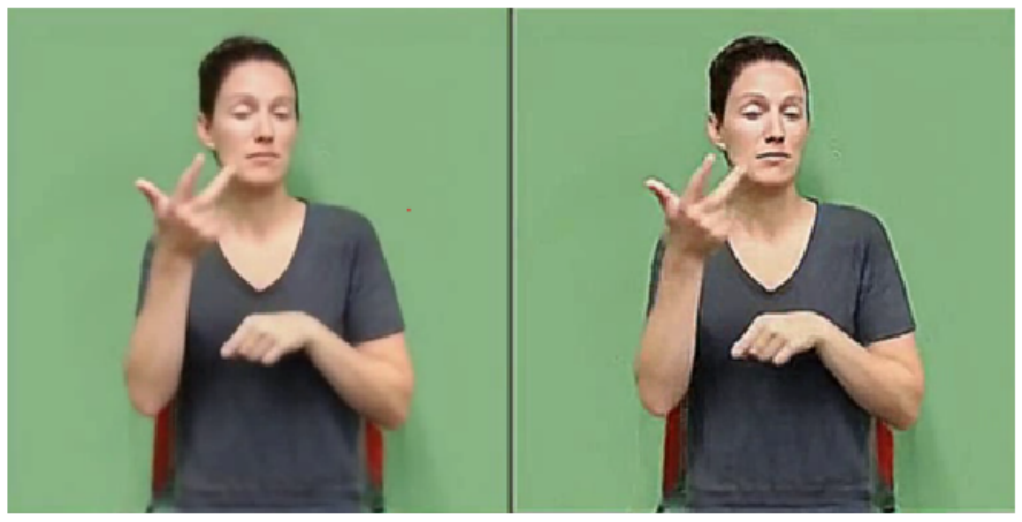

Fig. 5. A comparison of our raw model output on the left, versus the sharpened output on the right

*Post-processing*

Further enhancements for the face and hands was achieved by using a conventional sharpening technique. The filter used resembles a Laplacian filter, this was thenFollowing this, we applied the followed by a Gaussian filter [35] for smoothing. This resulted in the best possible generated videos that we could obtain, given our limited computational resources. The details of the fingers and facial features are much clearer and pronounced compared to those without the sharpening.

Figure 5 shows how big of a difference this sharpening actually makes to the final result. Visually, it is quite substantial.

For generating deepfakes of completely unseen subjects (32 subjects entirely outside of the real dataset), we incorporated SimSwap [5] (a face swapping approach) for better facial identity transfer. This was due to the current model not performing as well for such subjects.

## VI. Techniques to Analyze Sign Language Deepfakes

The analysis of the obtained deepfake videos includes a dual perspective: the visual aspect and the technical aspect. The visual aspect involves training machine learning algorithms to establish a baseline performance on the dataset. In contrast, the technical aspect evaluates the quality of the produced sign language interpretation relative to the original video, aiming to balance visual appeal and linguistic accuracy. This approach aims to measure the credibility of the deepfake dataset.

### A. *Linguistic Analysis*

We need to make sure that generated deepfake videos are linguistically plausible and understandable. Hence, for this stage of our work, we requested transcription and labeling of the videos from a sign language expert, Mr. Carlos Ivan Colon, at the University of Illinois at Urbana-Champaign, IL, USA.

After obtaining the transcriptions, we compared the ones in the real videos and the generated deepfake videos. Our analysis encompassed a variety of approaches, and employed a diverse range of techniques to ensure robustness of our results. Various employed comparison techniques are summarized in Table I.

*1) Experimental Setup for Linguistic Analysis:* As mentioned earlier, we sought help from a sign language expert to transcribe the real and generated sign language deepfake videos. The expert was not aware as to which videos were real and which were fake. We extracted 33 random real videos from the dataset. Subsequently, 2 generated deepfake videos were randomly selected for each real driving video, resulting in a pool of 66 fake videos. The sign language expert transcribed each video and also labeled whether the transcribed video was real or fake. Sharpening was applied to both real and fake videos before model training and testing to maintain fairness. The primary objective here was to assess the following:

- whether the generated deepfake videos linguistically mean the same as the real driving video
- whether the sign language expert finds it difficult to identify the real from fake videos

### B. *Visual Analysis - Establishing a baseline*

For establishing the baseline performance on the generated dataset, we used the following algorithms. Two of these are standard Machine Learning algorithms, while the other two are Deep Learning ones. One of them takes the temporal aspect into account as well.

1) ConvLSTM (Convolutional Long Short Term Memory)
2) CNN (Convolutional Neural Network)
3) RF (Random Forest)
4) SVM (Support Vector Machine)

*1) Experimental Setup for Visual Analysis:* The study employed two different scenarios. In one scenario, the subject identities in the training and testing set were completely different, while in the other scenario same driving videos of different subjects could appear during training and testing. However, the apparent subject identities were still different in training and testing. For each of these scenarios, we divided the data in three subsets.

The subset of each scenario included 150 real and 150 fake videos in the train set, with 50 real and 50 fake videos each in the test set. The video resolution was then down-sampled from 384x384 to 96x96 due to computational constraints. The dataset comprised of randomly selected fake videos from our deepfake dataset, while real videos were randomly drawn from the How2Sign dataset. Both real and fake videos underwent sharpening before being utilized in training and testing, ensuring a fair experiment.

It is important to mention that, excluding ConvLSTM, the remaining three models were assessed on 100 frames from each video. When more than 50% of the frames were labeled as real during testing, the entire video would be labeled as real. This decision was necessary as these models did not consider temporal context. Additionally, given the complexity of deep learning models, we incorporated a validation set to prevent overfitting. Each model maintained consistent parameters, train, validation, and test set sizes throughout training on each fold to ensure fairness.

## VII. Results & Discussion

We present the results and related discussion in the following sections. Each of these is equally important in showing how far we have come in terms of generating sign language production.

### A. *The Deepfake Dataset*

The deepfake dataset comprised 1212 videos, encompassing a total of 38 subject identities. Note that 32 subjects were entirely unseen and not part of the original dataset, and were introduced to ensure a fair analysis of model performance, particularly in the context of sign language. The dataset included videos with the 3 distinct types or combinations of style and pose transfer, as previously mentioned. Many of these videos exhibited high quality, but, as anticipated, those featuring completely unseen subjects tended to exhibit more artifacts. In Figure 6, we can see the results of some frames taken from our dataset. Visually, all 3 types of style and pose transfer proved to be quite appealing, specially when considering the resources required for model training and input image size to the model. Complete information on the dataset can be viewed in Table II.

### B. *Expert Opinion*

When it comes to the sign language interpreter and his analysis, we can make clear conclusions after looking at the confusion matrix in Figure 7 and the values in Table III. The interpreter exhibited an element of confusion in correctly classifying the real and fake videos since the accuracy is under 85%. However, the expert had more difficulty in detecting fake videos and mislabeled a number of them. This can be deduced from the confusion matrix in Figure 7 and from the sensitivity and specificity values. It is evident that the expert find it harder to spot fake videos.

### C. *Linguistic Analysis*

These results are intriguing. Each similarity metric, ranging from the most stringent to the most lenient, consistently meets the criteria. The transcripts of the fake videos align closely with those of the real videos, as evidenced by the distribution plots in Figure 8 and box plots in Figure 9. Most scores exhibit higher medians and narrower inter-quartile ranges.

TABLE I
SUMMARY OF LINGUISTIC ANALYSIS TECHNIQUES

| **Technique** | **Purpose** | **Score Range** |
|---|---|---|
| BLEU (Bilingual Evaluation Understudy) | Evaluate machine-generated text against human reference text | 0 to 1 (higher values indicate better performance) |
| Jaccard Similarity | Assess diversity and similarity of two sets of words | 0 to 1 (higher values indicate greater similarity) |
| Cosine Similarity | Measure similarity in the direction of vectors representing texts | 0 to 1 (higher values indicate greater similarity) |
| Levenshtein Distance (Edit Distance) | Quantify the minimum modifications needed to align two strings | Lower values indicate greater similarity |
| ROUGE (Recall-Oriented Understudy for Gisting Evaluation) | Evaluate overlap in n-grams between generated and reference text | 0 to 1 (higher values indicate better performance) |
| Jaro-Winkler Similarity | Compare sequences, emphasizing phonetic and prefix similarity | 0 to 1 (higher values indicate greater similarity) |

TABLE II
DEEPFAKE DATASET STATISTICS

| **Statistic** | **Value** |
|---|---|
| Total Deepfake Videos | 1212 |
| Female Videos | 560 |
| Male Videos | 652 |
| Average Duration (seconds) | 8.67 |
| Total Subjects | 38 |
| Subjects in Original Dataset | 6 |
| Completely Unseen Subjects | 32 |

TABLE III
PERFORMANCE METRICS OF INTERPRETER

| Specificity | Sensitivity | Accuracy |
|---|---|---|
| 0.9091 | 0.8030 | 0.8384 |

It is worth noting that BLEU doesn't seem to perform as well. However, given its strict nature, the best scores are usually around 0.6 to 0.7 although the maximum limit is 1, and our median lies in this range. The reason for this high strictness is that the BLEU score looks for an exact match between two sentences. Similarly when it comes to other scores, they too approach their desirable or acceptable values according to the natural language processing literature [36], [37], [38], [39], [40], [41].

This observation is supported by the distribution plots in Figure 8, where the highest frequency is at score 1 for relevant models. Similarly, it is 0 for Levenshtein distance and Jaccard similarity. One can note that most of the histograms are concentrated close to their optimum value.

Examining the box plots in Figure 9 reveals another interesting finding. When computing different types of ROGUE scores, the medians consistently remain high, ranging roughly from 0.75 to 0.85. This is significant because in ROGUE 2, which is the strictest ROGUE score, the median falls to about 0.75 and what makes it strict is that it looks at the ordering of a greater number of ordered bigrams. This means our fake video transcripts continue to perform exceptionally well even for the worst ROGUE score.

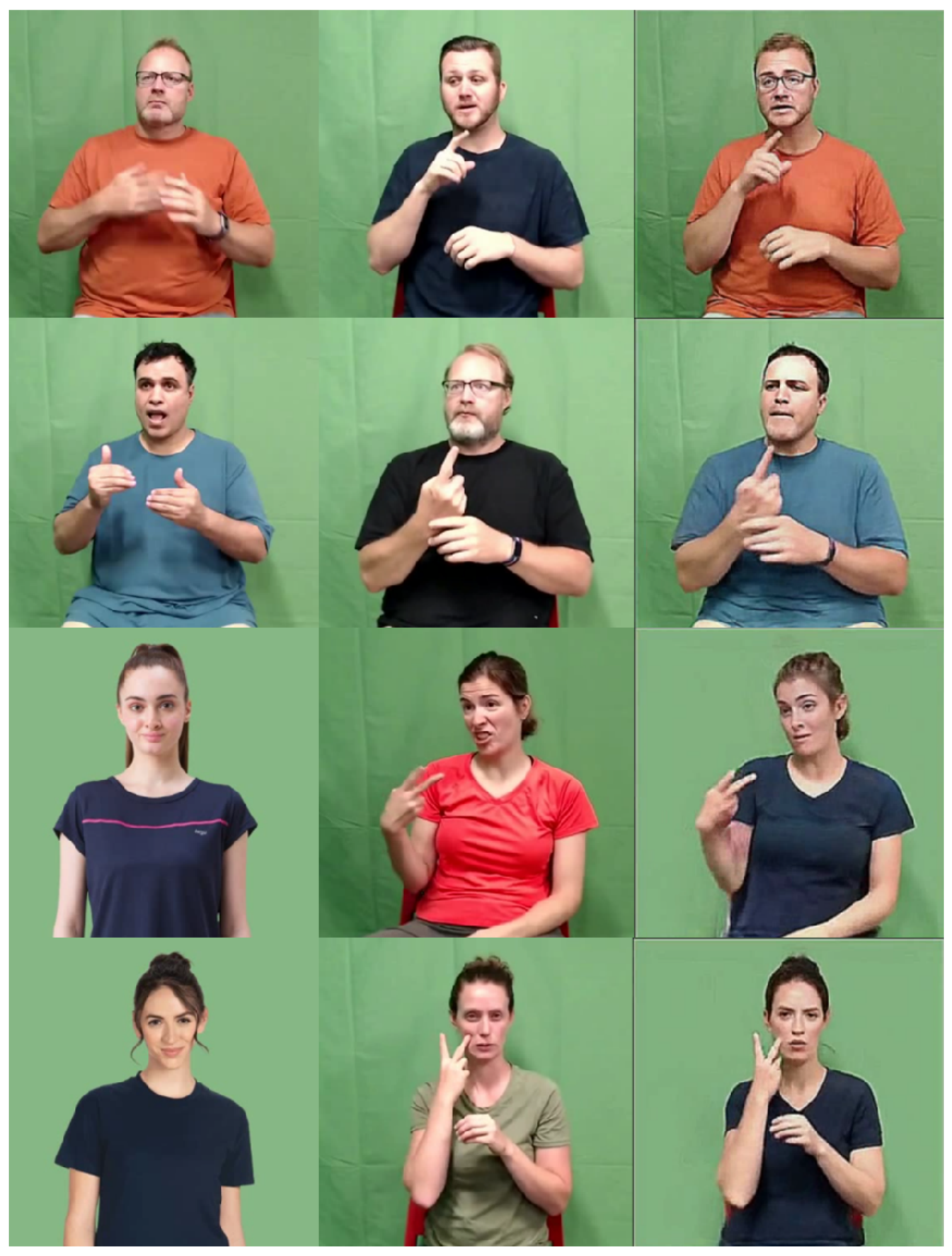

Fig. 6. Some examples of our deepfake dataset. Here we have the source image on the left, the driving video frame in the middle and the output video frame on the right

Henceforth, the above suggests that our transcripts not only exhibit similarity in terms of meaning but also in the precise pattern of words. Therefore, considering the linguistic and technical aspects discussed earlier, our fake videos can be deemed credible and acceptable.

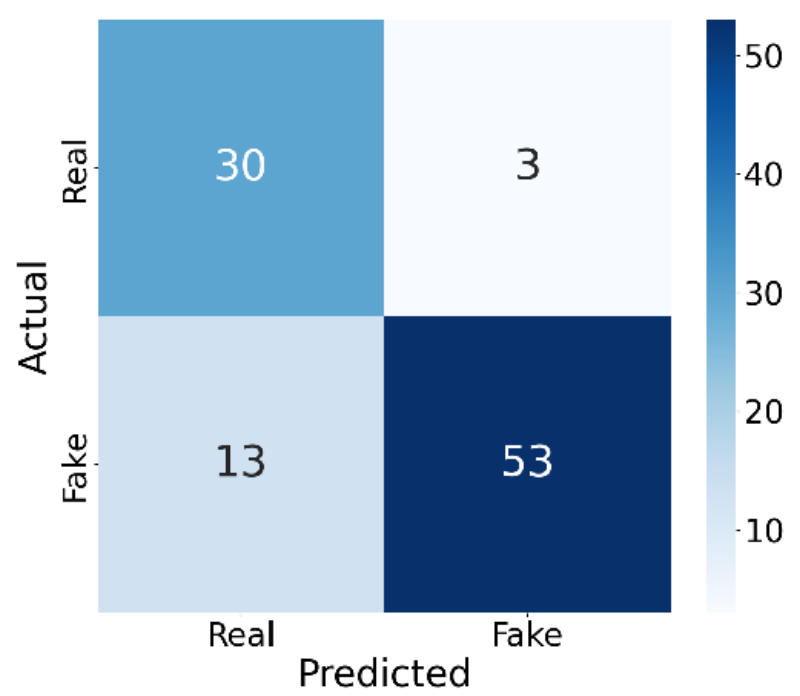

Fig. 7. Confusion matrix for video classification by interpreter

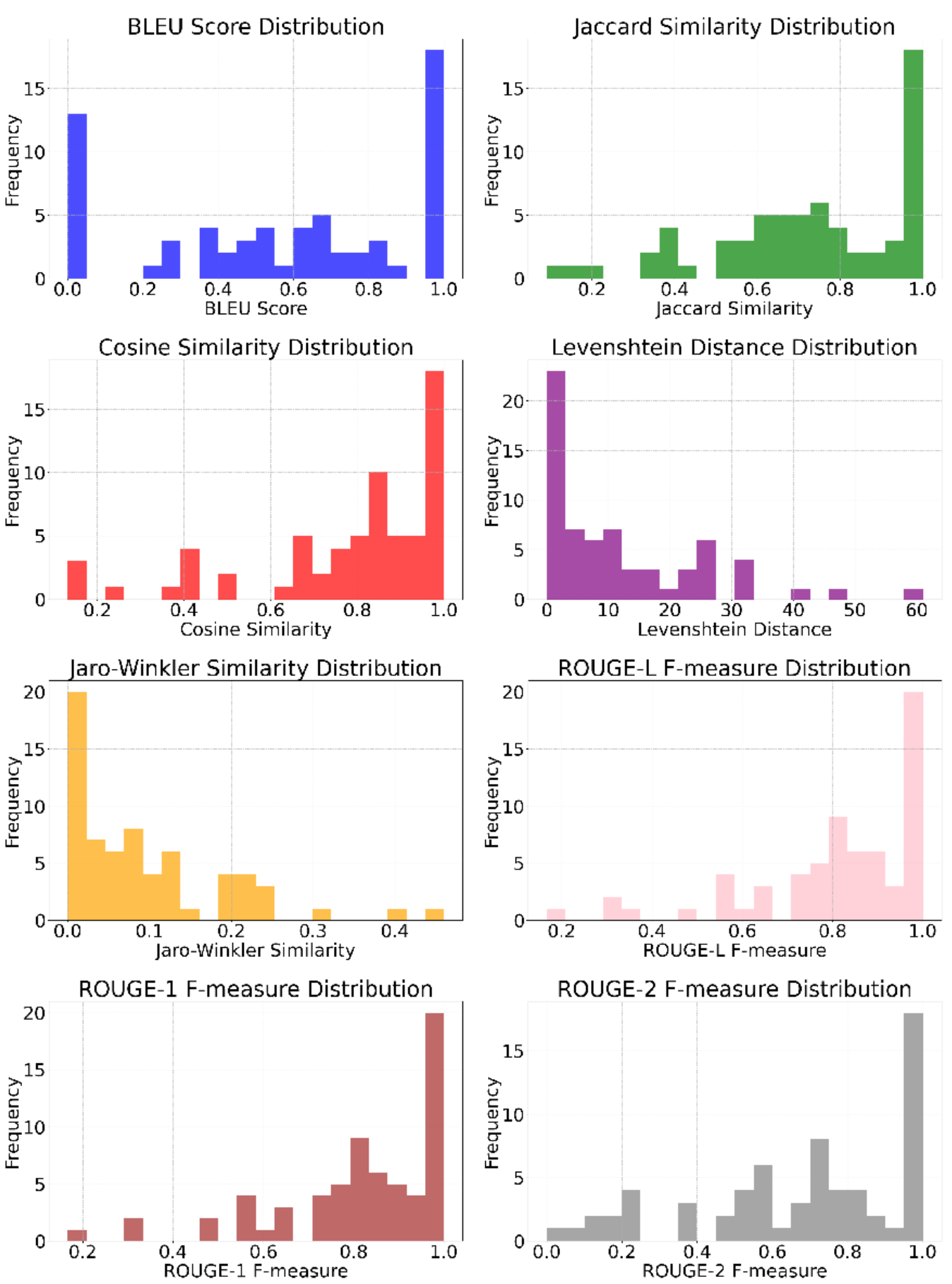

Fig. 8. Distribution plots for the various similarity scores

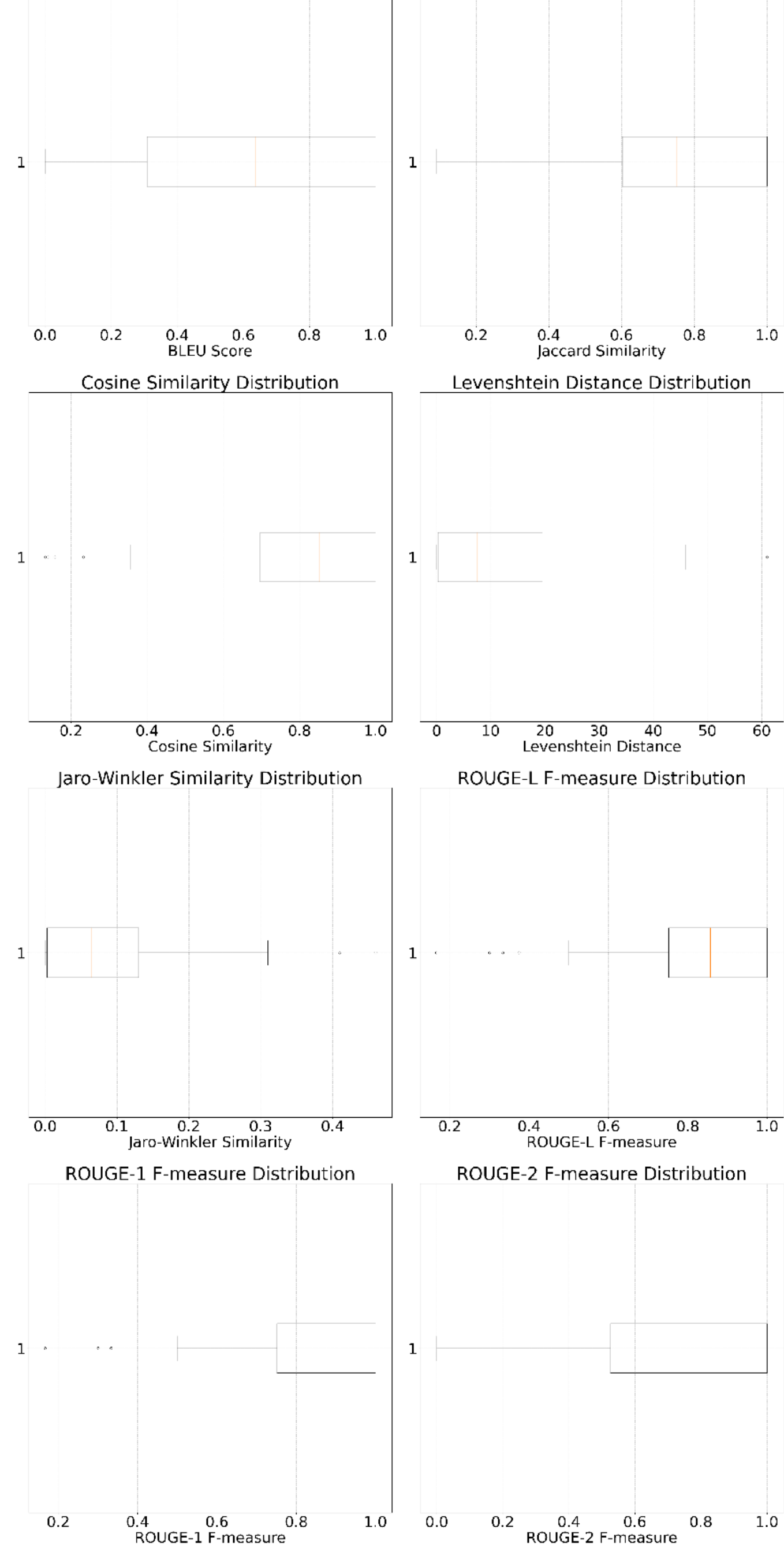

Fig. 9. Box plots for the various similarity scores

### D. Visual Analysis

As for these results, the first thing one can note after seeing the performance metrics bar plots in Figure 10 is that the accuracy increases as we move from the independent folds to the sub-independent folds except in SVM. This makes sense since there is a greater overlap between the train and test sets of the sub-independent folds. In terms of average accuracy across folds, RF performs the best.

The examination of both the confusion matrices in Figure 11 and the performance metrics plots in Figure 10 reveals an interesting trend across all models. When transitioning from independent folds to sub-independent ones, we observe significant changes in both sensitivity and specificity, except for the RF model. This trend aligns with the variations in information overlap across different fold configurations.

Notably, RF consistently demonstrates the best specificity, while SVM surprisingly shows better sensitivity despite having lower accuracy. This means RF is great at spotting fake videos, while SVM is good at recognizing genuine ones, even though its overall accuracy is lower in our testing. However, this is not necessary as the decision thresholds can be modified to change this but after experimenting with various parameters we got the highest overall accuracy with these thresholds. Therefore, keeping this is mind, our conclusion in this regard stands true.

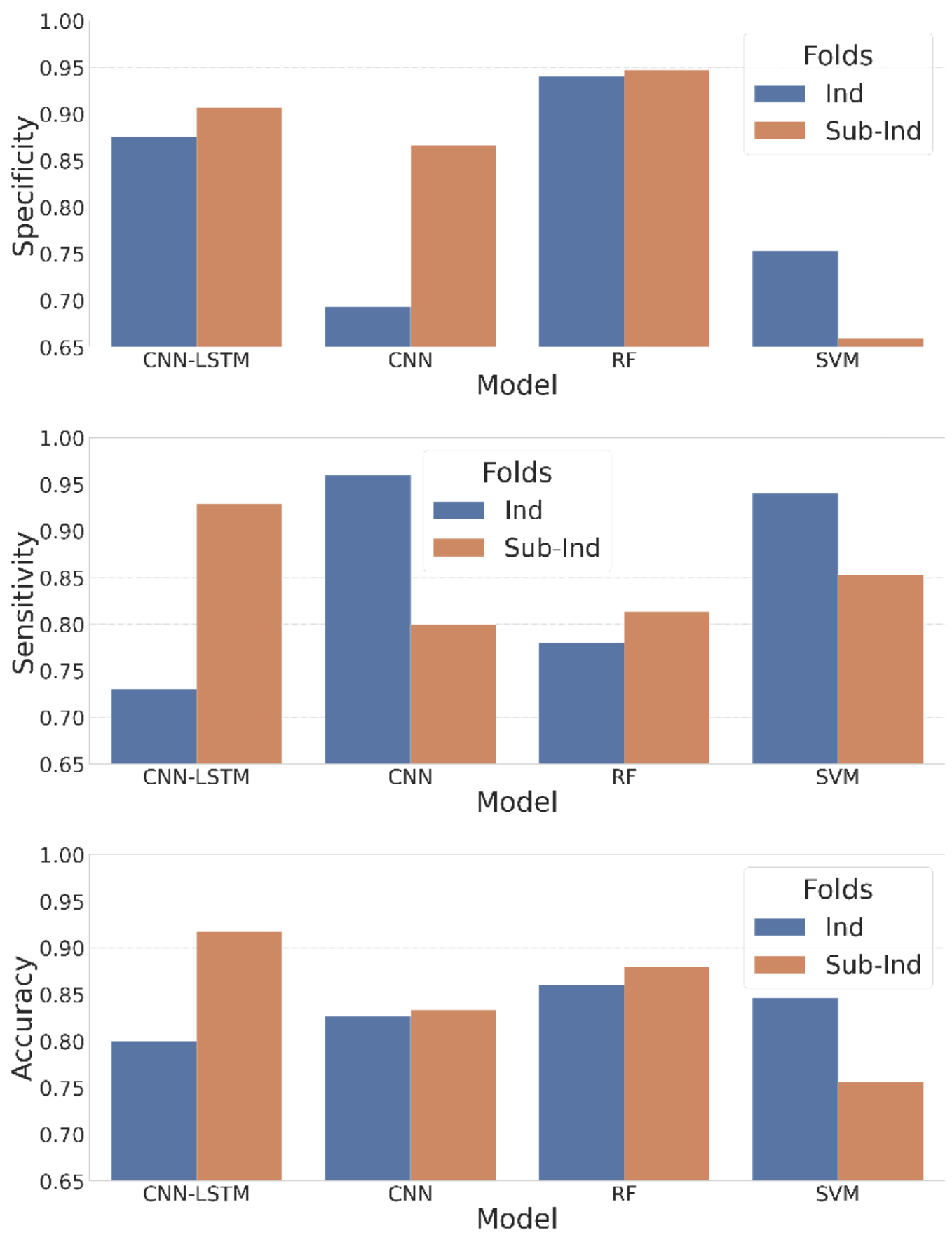

Fig. 10. Bar graphs for the various performance metrics of the mode/scenario combinations

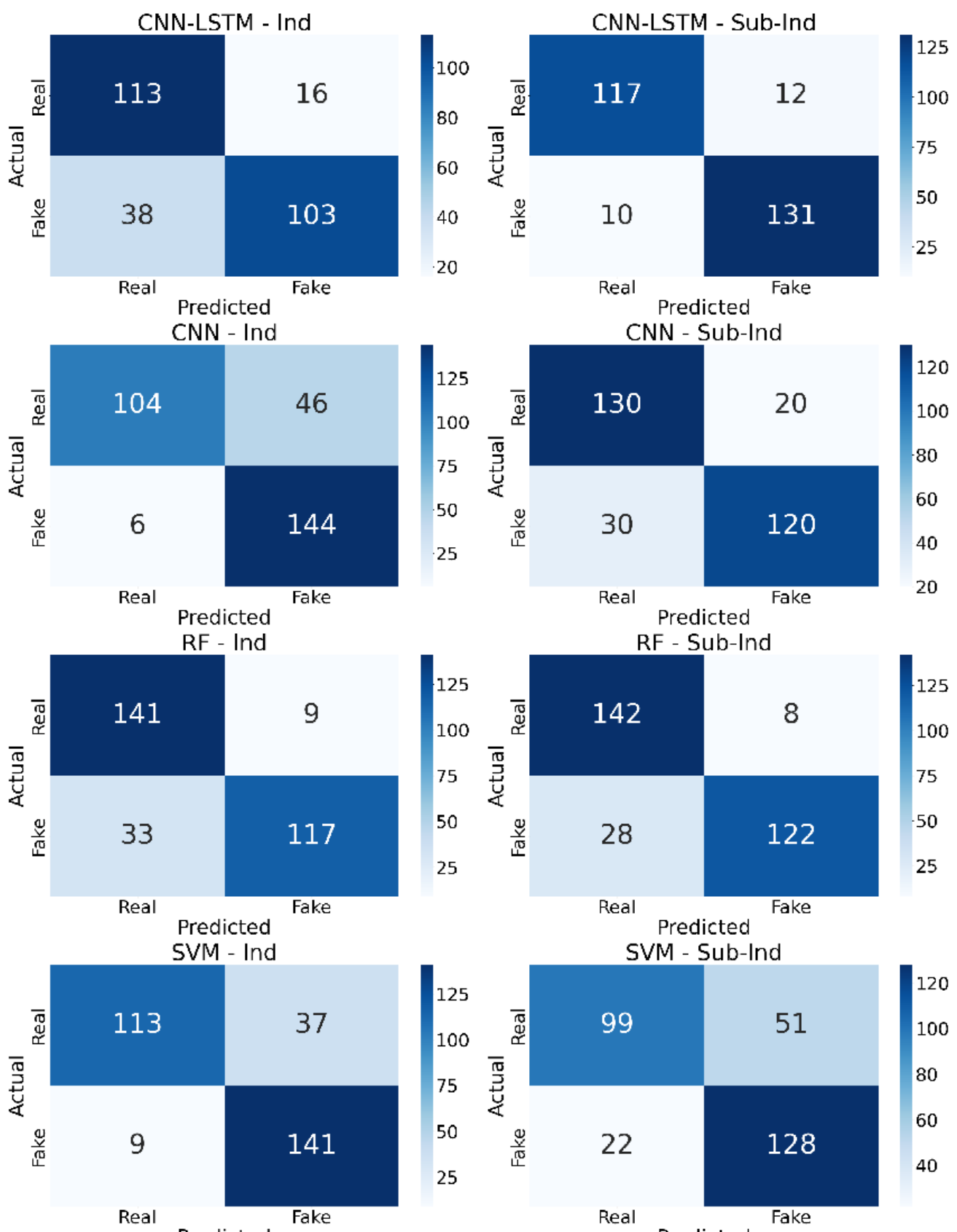

Fig. 11. Confusion matrices for the various model/scenario combinations

The strong performance of RF is highlighted by its ability to remain consistent across the two fold types, as seen in its confusion matrices across folds in Figure 11. This consistency is unusual, as none of the other models exhibit this behavior.

Another interesting observation is the increase in specificity for all models, except SVM, when moving from independent to sub-independent folds. This might be due to better alignment between the features of the training and test sets in the sub-independent folds.

However, our analysis underscores the intentional design of a sign language dataset aimed at inducing confusion visually. Models excelling at classifying fake videos often struggle with confusion in accurately classifying real videos, and vice versa. This suggests that our deepfake videos possess enough realistic visual elements, posing a challenge for machine learning algorithms to precisely classify every video. These results can serve as baseline for the future researchers.

## VIII. Conclusion

This research successfully generated a large dataset of over 1200 sign language deepfake videos that are visually and linguistically credible. The videos showcase diverse style and pose transfers between real subjects. Rigorous analysis verified the authenticity and interpretability of the fake videos. Machine learning models struggled to perfectly classify real versus fake, demonstrating intentional visual realism. A sign language expert also exhibited confusion in identifying deepfake sign language videos. The work establishes a robust benchmark for developing deepfake detection in sign languages. By producing credible synthetic videos, the dataset will drive future innovation in exposing fake sign language content aimed at spreading misinformation. The consistent high similarity scores between real and fake video transcripts firmly validate the interpretability of the generated content. The research makes major strides in ethical deepfake production to equip the community with assets for combating fake news targeting vulnerable groups. In addition to advancing sign language video generation, this work lays the foundation for critical deepfake detection research.

## Acknowledgment

This work has been funded in-part by the grant FRG21-M-E94 from the American University of Sharjah.

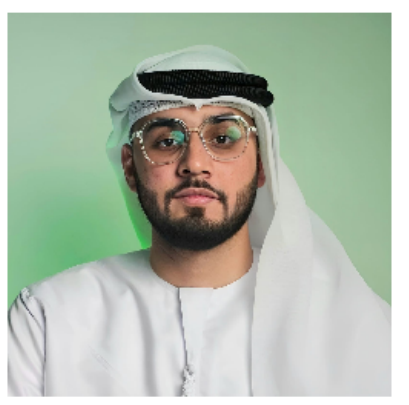

**Shahzeb Naeem** is a Masters student at the American University of Sharjah. He earned his Bachelors from the National University of Sciences and Technology, Islamabad. Before joining AUS, he worked as an AI Engineer at a startup while working on cutting-edge tech related to object detection, tracking, and generative AI. Currently, he is working as an AI Researcher along with his studies at his university.

His research interests include the previous topics as well as computer vision, image processing, and machine learning in general. He has won competitions such as the UAE and Arab IOT and AI Challenges and has been rewarded the CureMD Young Inventor and their ForteQuest Champion among other achievements.

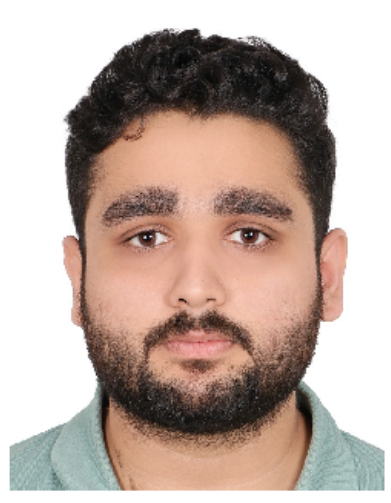

**Muhammad Riyyan Khan** was a Masters student at the American University of Sharjah. He too earned his Bachelors from the National University of Sciences and Technology, Islamabad. Before joining AUS, he worked as a Software Engineer at a startup while working the latest technology related to web development and an eventual shift to AI. Currently, he is working as an AI Researcher along at his university.

His research interests also include computer vision, image processing, and machine learning in general. He has won competitions such as the UAE and Arab IOT and AI Challenges and has been rewarded Arcelik Global (Dawlance) Top Intern.

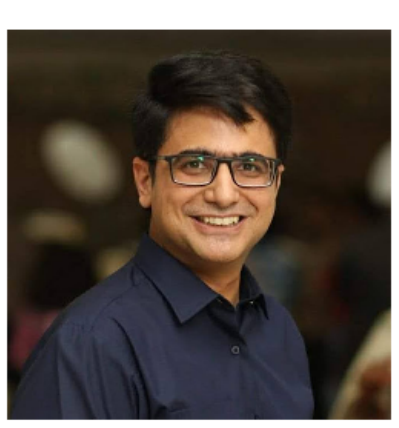

**Usman Tariq** is an Associate Professor in the Department of Electrical Engineering at the American University of Sharjah. He earned his PhD and MS in Electrical and Computer Engineering from the University of Illinois at Urbana-Champaign. Before joining AUS, he worked as a research scientist at the Xerox Research Center in Europe. His research interests include computer vision, image processing, and machine learning, with a focus on deepfakes and physiological signals.

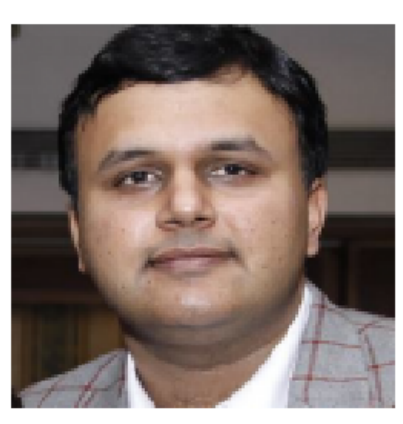

**Abhinav Dhall** is an Associate Professor at the Data Science and AI Department, Monash University, Australia. Before joining Monash University, he led the Centre for Applied Research in Data Sciences at the Indian Institute of Technology Ropar, where he introduced a joint Masters program and an industry-centric certification program in data sciences. At Monash University, he has also co-directed the Human-Centred Artificial Intelligence Lab.

He received his PhD from the Australian National University, supported by the Australian Leadership Award. Additionally, he was a visiting scholar at the University of California San Diego and Imperial College London. He completed postdoctoral fellowships at the University of Waterloo and the University of Canberra.

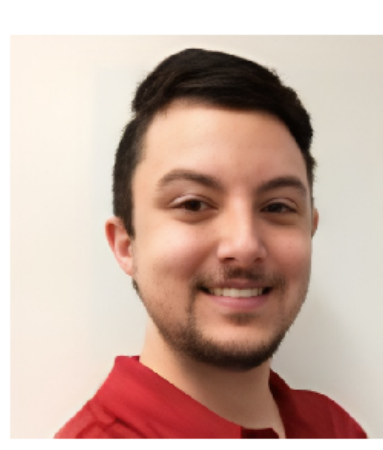

**Carlos Ivan Colon** is a full-time faculty member teaching American Sign Language at the University of Illinois at Urbana-Champaign since August 2019. Previously, he worked as an American Sign Language instructor at two other universities after graduating with a Bachelor of Arts from Gallaudet University in May 2017.

In December 2019, he earned a Master of Arts in Teaching American Sign Language from the University of Northern Colorado. His expertise includes developing innovative approaches to teaching American Sign Language and fostering inclusivity for Deaf and hard-of-hearing communities.

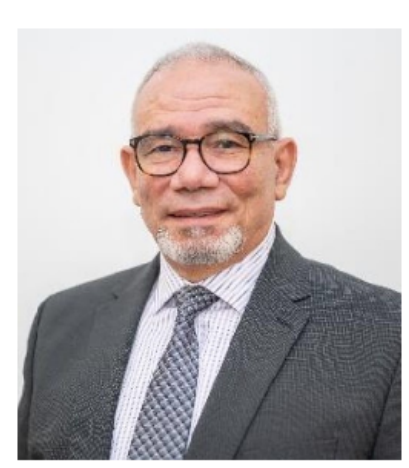

**Hasan Al Nashash** is a Professor at the American University of Sharjah. He coordinates the PhD program in Biosciences and Bioengineering and was the Director of the Biomedical Engineering MS Program and the Department of Electrical Engineering. His research interests include mental stress management, brain source localization, spinal cord and brain injuries, and low-power electronic devices.

He has authored over 150 papers, six book chapters, and holds two US patents. He is the Vice Chair of the EMB Standards Committee (EMB-SC) and has received awards including the Roderick French Distinguished Service Award. He collaborates with institutions like the National University of Singapore, Johns Hopkins University, and Rashid Hospital in Dubai.